\documentclass{article}
\pdfoutput=1

\PassOptionsToPackage{numbers,compress}{natbib}
\usepackage[eandd, preprint]{neurips_2026}

\usepackage{graphicx}
\usepackage{booktabs}
\usepackage{sidecap}
\usepackage{amsfonts}
\usepackage{amsmath}
\usepackage{nicefrac}
\usepackage{microtype}
\usepackage{xcolor}
\usepackage{wrapfig}
\usepackage{float}
\usepackage[font=small]{caption}

\usepackage[english]{babel}
\usepackage{rotating}
\usepackage{gensymb}
\usepackage{relsize}
\usepackage{makecell}

\usepackage{listings}

\usepackage{longtable}

\lstdefinestyle{pythonstyle}{
    language=Python,
    basicstyle=\ttfamily\small,
    numbers=left,
    numberstyle=\tiny,
    breaklines=true,
    keywordstyle=\color{blue},
    commentstyle=\color{gray},
    stringstyle=\color{teal},
}

\definecolor{turquoise}{cmyk}{0.65,0,0.1,0.3}
\definecolor{purple}{rgb}{0.65,0,0.65}
\definecolor{dark_green}{rgb}{0, 0.5, 0}
\definecolor{orange}{rgb}{0.8, 0.6, 0.2}
\definecolor{red}{rgb}{0.8, 0.2, 0.2}
\definecolor{darkred}{rgb}{0.6, 0.1, 0.05}
\definecolor{blueish}{rgb}{0.3, 0.3, .6}
\definecolor{light_gray}{rgb}{0.7, 0.7, .7}
\definecolor{pink}{rgb}{1, 0, 1}
\definecolor{greyblue}{rgb}{0.25, 0.25, 1}
\definecolor{awesome}{rgb}{1.0, 0.13, 0.32}

\definecolor{figscene}{HTML}{2A78D6}
\definecolor{figobject}{HTML}{EB6834}
\definecolor{figmaterial}{HTML}{1BAF7A}

\definecolor{figred}{rgb}{0.9, 0.1, 0.1}
\definecolor{figgreen}{rgb}{0.1, 0.7, 0.1}
\definecolor{figblue}{rgb}{0.1, 0.1, 0.9}
\definecolor{figmagenta}{rgb}{0.8, 0.1, 0.8}

\usepackage{arydshln}

\makeatletter
\renewcommand\paragraph{\@startsection{paragraph}{4}{\z@}%
	{0ex \@plus.3ex \@minus.2ex}%
	{-1em}%
	{\normalfont\normalsize\bfseries\maybe@addperiod}}
\newcommand{\maybe@addperiod}[1]{#1\@addpunct{.}}
\makeatother

\newcommand{\datasetname}{{New York Smells}}

\usepackage[breaklinks,colorlinks,citecolor=blue]{hyperref}

\title{New York Smells:\\A Large Multimodal Dataset for Olfaction}

\author{%
  Ege Ozguroglu\textsuperscript{1} \quad
  Junbang Liang\textsuperscript{1} \quad
  Ruoshi Liu\textsuperscript{1} \quad
  Mia Chiquier\textsuperscript{1} \quad
  Michael DeTienne\textsuperscript{2} \\ \textbf{
  Wesley Wei Qian\textsuperscript{2} \quad
  Alexandra Horowitz\textsuperscript{1} \quad
  Andrew Owens\textsuperscript{3} \quad
  Carl Vondrick\textsuperscript{1}} \\
  {\normalfont\textsuperscript{1}Columbia University \quad
  \textsuperscript{2}Osmo Labs \quad
  \textsuperscript{3}Cornell University}\\
  \url{http://smell.cs.columbia.edu}
}

\begin{document}

\makeatletter
\let\@notice\relax
\makeatother

\maketitle

\begin{abstract}

While olfaction is central to how animals perceive the world, this rich chemical sensory modality remains
largely inaccessible to machines. One key bottleneck is the lack of diverse, multimodal olfactory training data
collected in natural settings. We present {\em \datasetname}, an in-the-wild dataset of paired
image and olfactory signals. Our dataset contains 7,000 smell-image pairs from 3,500 %
distinct objects across indoor and outdoor environments, with approximately 70$\times$ more objects than existing olfactory
datasets. Our benchmark
measures cross-modal smell-to-image retrieval, and recognition of objects and materials from smell alone. Through experiments on our dataset, we find that visual data enables cross-modal olfactory representation learning, and that our learned  olfactory representations outperform widely-used hand-crafted features.

\end{abstract}

\section{Introduction}

\begin{figure}[!b]
\centering
\includegraphics[width=\linewidth]{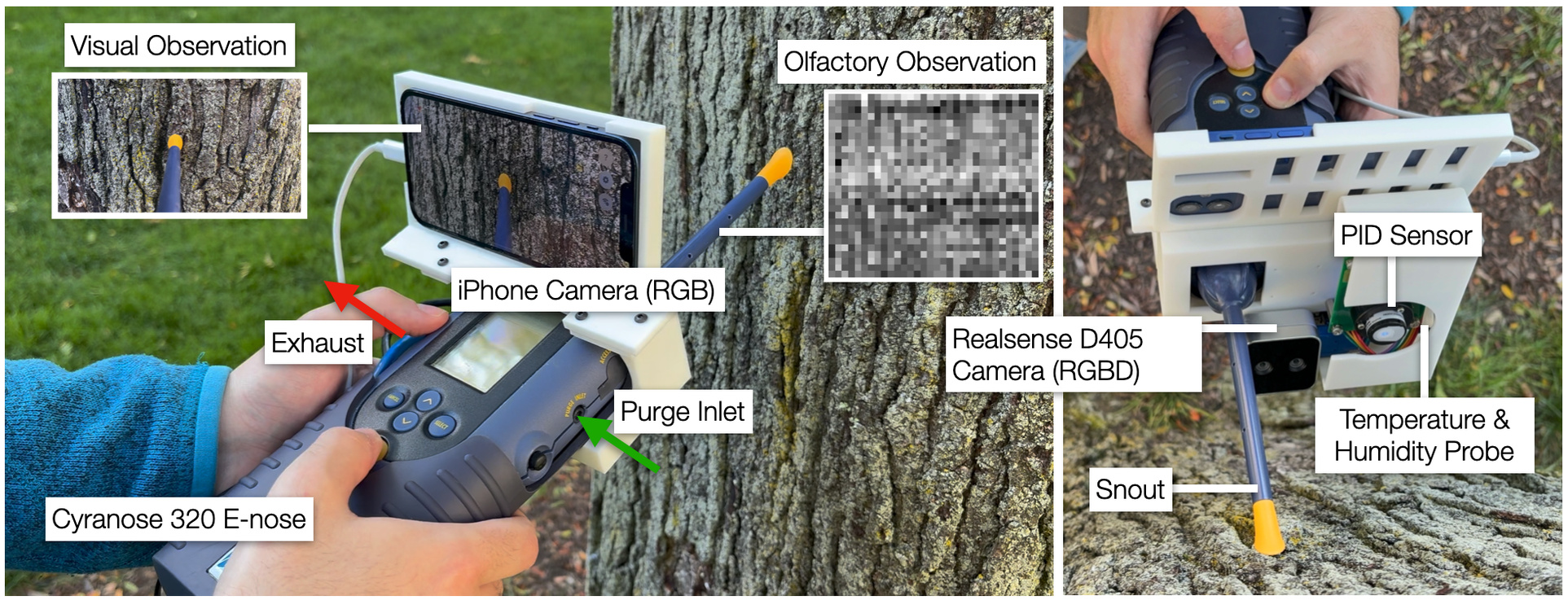}
\caption{{\bf Linking Vision and Olfaction:} We introduce a multimodal dataset of real-world scenes, capturing paired olfaction and visual signals using a camera mounted to an e-nose on a custom 3D-printed sensor rig. We also capture a suite of other supplementary modalities: depth (from an RGB-D camera), temperature, humidity, and ambient VOC concentrations (from a PID sensor). } %
\label{fig:capture_setup}
\end{figure}

Olfaction---the sense of smell---is a key way that animals, and to a lesser extent humans, perceive the world. Yet, this rich ``chemical world'', central to the sensory experience of many species, is largely imperceptible to machines. This is in contrast to sight, sound, and touch, where advances in machine learning, particularly unsupervised and multimodal learning, have led to rapid improvements in machine abilities. One of the major obstacles to applying this approach to olfaction is the lack of suitable data. Existing olfaction datasets have largely been based on perceptual descriptors, rather than the raw outputs of olfactory sensors, or are captured in lab settings. Unlike audio and touch~\cite{chen2020vggsound,yang2022touch,girdhar2023imagebind}, existing olfactory datasets are not paired with vision (or other sensory modalities), making it difficult to link olfaction to the representations of other modalities.

In this paper, we introduce {\em {\datasetname}}, a large in-the-wild dataset of paired vision and olfaction. We visited dozens of indoor and outdoor scenes, such as parks, gyms, dining halls, libraries, and streets. We walked through each scene and recorded naturally synchronized images and smells of their odorant objects. To supplement this data, we also record a suite of other sensors: RGB-D, temperature, humidity, and volatile organic compound (VOC) concentration. Our dataset, which contains $7000$ olfactory-visual samples from $3500$ objects, is significantly larger and more diverse than other olfactory datasets. It contains $70 \times$ as many distinct objects as the contemporaneous {\em lab-collected}, single-modality dataset of Feng et al.~\cite{Feng2025}, the most similar work to ours.  %

We use our dataset to establish a benchmark for in-the-wild and multimodal smell perception. We propose a smell-to-image retrieval task that evaluates the ability of a model to establish cross-modal visual-olfactory associations. We also evaluate recognition of objects and materials from odors alone, using vision as the supervisor only during training.
To further demonstrate the utility of paired visual-olfactory signals, we use our dataset for multimodal representation learning. Inspired by self-supervised learning methods in other multimodal domains~\cite{arandjelovic2017look,tian2020contrastivemultiviewcoding,radford2021learningtransferablevisualmodels}, we train general-purpose olfactory features by training a joint embedding between smell and sight using contrastive learning. Through experiments on our downstream tasks from our benchmark using a variety of different network architectures, we find that the olfactory representations learned using our dataset significantly outperform hand-crafted smell features that are widely used in prior work.

\begin{figure}[t]
\centering\includegraphics[width=\linewidth]{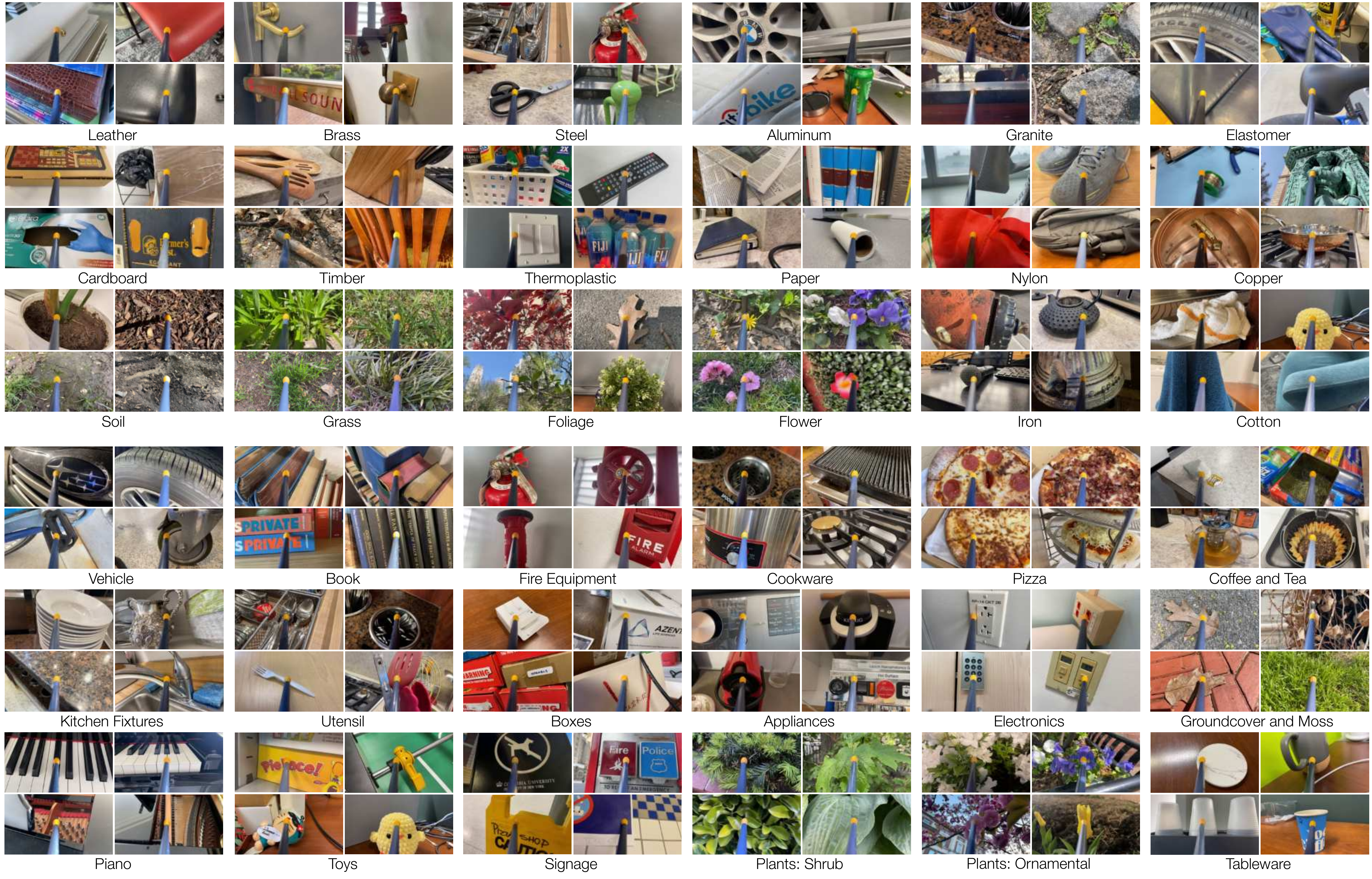}
    \caption{\textbf{Dataset Gallery:} We collected a diverse dataset of paired sight and olfaction by visiting many locations within New York City and recorded a variety of materials (top rows) and objects (bottom rows) in different scenes. We show a selection of the captured images here. All samples have a corresponding olfactory signal captured from the Cyranose electronic nose.}
    \label{fig:dataset}
\end{figure}

\begin{figure}
    \centering
    \includegraphics[width=\linewidth]{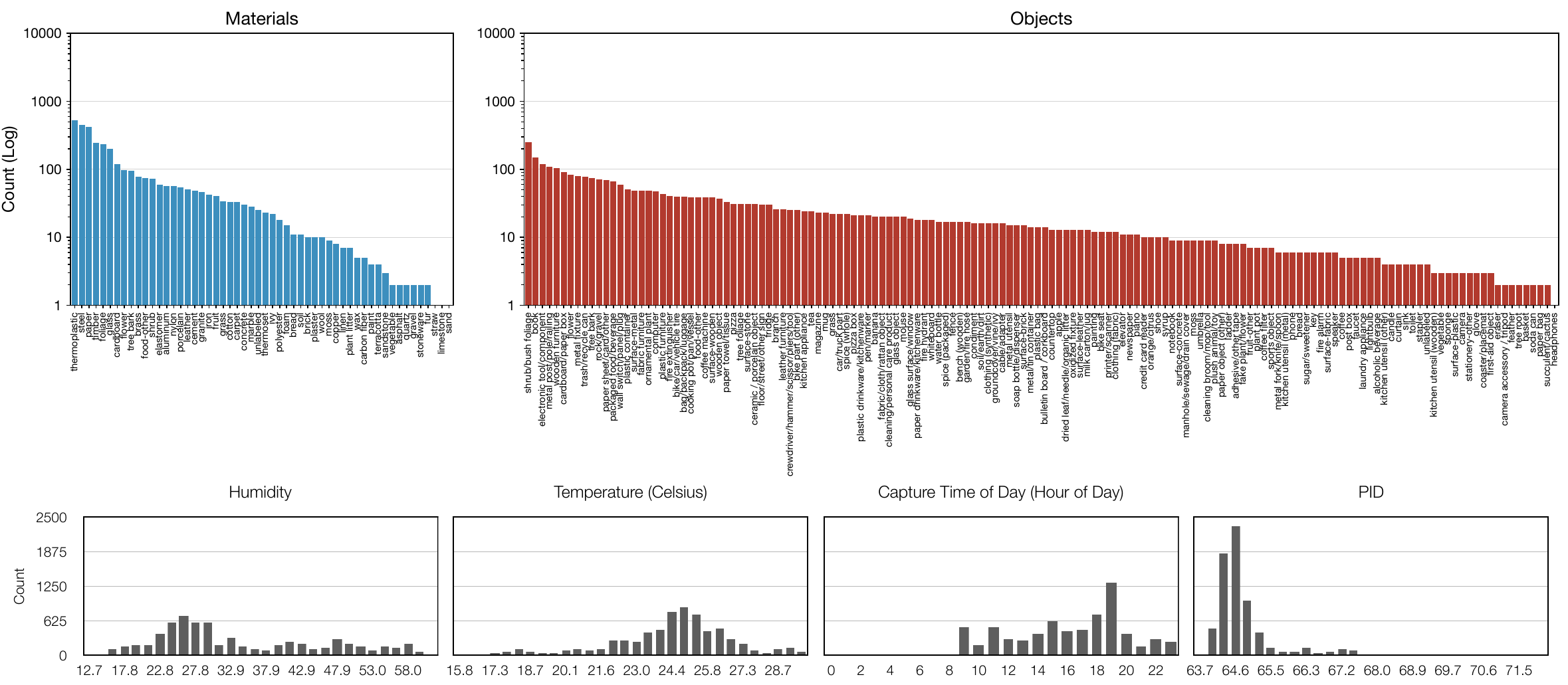}
    \caption{{\bf Dataset Statistics.} Our in-the-wild dataset covers a wide variety of objects, materials, and ambient conditions.}
    \label{fig:histogram}
\end{figure}

We see this dataset as a step toward in-the-wild, multimodal olfactory perception, as well as a step toward linking sight with smell.  While olfaction has traditionally been approached in constrained settings, such as quality assurance, there are many applications in natural settings. For example, as humans, we constantly use our sense of smell to assess the quality of food, identify hazards, and detect unseen objects. Moreover, many animals, such as dogs, bears, and mice, show superhuman olfaction capabilities \cite{kokocinska2021canine}, suggesting that human smell perception is far from the limit of machine abilities. 
\noindent Our work makes the following contributions:
\begin{itemize}
\item Our dataset provides much more diverse and naturalistic olfaction recordings than previous datasets.
\item Our dataset is the first to pair in-the-wild olfaction with images.
\item Using the visual signals in our dataset, we establish an evaluation for in-the-wild olfactory understanding.
\item We show that visual signals from our dataset provide supervision for general-purpose olfactory feature learning.
\end{itemize}

\section{Related Work} \label{sec:related}

\paragraph{Machine Olfaction.}
Previous work in machine olfaction has often focused on idealized settings, often requiring a miniature chemistry lab to be embedded into the system, which is expensive, bulky, and often impractical. 
Research has studied processing molecular structures for the purpose of designing scents \cite{sanchez2019machine} by using graph neural networks on synthetic data to predict human preferences such as perfume, or how to modify crops to prevent pests from spreading \cite{fundurulic2023advances}.  Other work does recognition from real-world sensors, e.g.,\ detecting diseases like COVID \cite{ghazaly2023assessment} or explosive devices \cite{exkq-c076-19}.
Despite these advances, machine olfaction research has been limited to a) narrow domains that lack the diversity and realistic complexity of everyday situations, and b) heavily relied on exact molecular information, which requires hardware that is not available to portable, low-cost sensors. Our dataset instead is designed for diverse scenarios, where sensors are noisy and incomplete, and methods must scale to vast domains of in-the-wild scents and odors. %

Raw signals from electronic noses are high-dimensional and noisy, making data-driven methods attractive for uncovering structure. At the molecular level, psychophysical datasets enable models to predict perceptual attributes \cite{Keller2017}, and graph-based approaches propose a principal odor map (POM) \cite{Lee2023}. Mixture studies are limited, showing approximate perceptual similarity \cite{Snitz2013} and the existence of olfactory metamers \cite{Ravia2020}. Exploratory work has used mass spectra \cite{Debnath2020}, as well as ion-mobility and e-nose data \cite{Mueller2019}, but largely under lab conditions. Recent work emphasizes the importance of calibrating olfactory neuroscience to natural concentration ranges \cite{Wachowiak2025}, motivating the need for olfactory data in natural environments. In contemporaneous work, Feng et al.~\cite{Feng2025} collect a dataset of smells using an e-nose. However, their approach is limited to a highly controlled lab environment: to capture each example, they  place one object at a time in the same room. It is also relatively small scale, comprising 50 objects. By contrast, our work: 1) captures ``in-the-wild'' olfaction in natural environments of smells, 2) contains paired multimodal signals, 3) is much more extensive. We also go beyond prior work by using our dataset for multimodal representation learning.

We specifically focus on the Cyranose e-nose~\cite{sensigent_cyranose320}, since it is a popular, hand-held sensor that provides a rich olfactory signal that captures a variety of chemical properties. It has been applied to a range of scientific and industrial applications, such as measuring food quality~\cite{li2007neural,beghi2017electronic,zhou2021feasibility}, recognizing bacteria~\cite{balasubramanian2005identification,dutta2002bacteria}, evaluating the quality of construction materials~\cite{autelitano2019electronic}, detecting fires~\cite{ni2008orthogonal}, monitoring wildlife and fauna~\cite{doty2020assessment,ferreira2023first}, and disease detection~\cite{visser2020smell,shelley2025impact}.

\paragraph{Cross-Modal Supervision.}
There have been a variety of different methods for supervising one sensory modality using another. Early work by De Sa~\cite{de1993learning} proposed to use hearing to train vision through self-supervision. Ngiam et al.~\cite{ngiam2011multimodal} used a deep generative model to learn an audio-visual speech representation.  In contrast to these works, we use our dataset for {\em olfactory} representation learning through cross-modal supervision with sight. Our work is closely related to audio-visual~\cite{owens2016visually,aytar2016soundnet} and visual-tactile~\cite{yuan2017connecting, yang2022touch, dou2024tactile} data collection efforts in which a human probes objects with a sensor while recording video. By contrast, we pair olfaction with multiple visual sensors. The scale of our dataset is comparable to other in-the-wild paired-sensor efforts that share the same physical-proximity constraint---a human must bring the sensor into contact with each object---such as Touch and Go~\cite{yang2022touch} and ObjectFolder Real~\cite{gao2023objectfolder}. Recent work has learned a multimodal representation of taste~\cite{bender2023learning}. However, this approach is based solely on text descriptions of wine, whereas we use a real signal from a sensor.

\paragraph{Animal Olfaction.}\label{olfaction} This work is motivated in part by the olfactory capacity of animals. Domestic dogs in particular, are renowned for having an extraordinary sense of smell. Their ability is manifest in various detection tasks, identifying everything from the presence of bed bugs to landmines to owners' low blood sugar ~\cite{GadboisReeve2014}.
Anatomically, dogs have hundreds of millions more olfactory receptor cells (the cells that begin the translation of VOCs into the perception of an odor) than humans do~\cite{HepperWells2015Canidae}. This enables them to detect more smells and more types of smells at lower concentrations. Their noses have separate routes for smelling and respiration, which enables airflow to arrive at the olfactory epithelium with every inhale ~\cite{Craven2010}. The dog olfactory bulb is two percent of their brain by volume and sixty times the relative size of the human olfactory bulb~\cite{Hepper1988HumanOdorDog}.

\section{The \textit{\datasetname} Dataset}
\label{dataset}
We collect a large-scale dataset of natural olfactory-visual sensory data. Specifically, our dataset contains multimodal ``smell-centric" data. Unlike prior efforts on smell in machine perception \cite{Lee2023} and olfactory neuroscience \cite{Wachowiak2025}, or the contemporaneous SMELLNET dataset \cite{Feng2025}, which rely on controlled or synthetic environments and stimuli, our dataset is collected in-the-wild. We probe everyday objects in their natural environments using paired vision and olfaction sensors. This approach captures the range of naturally occurring odorant concentrations, a property that is key for modeling olfaction under natural conditions \cite{Wachowiak2025}. We will publicly release the full dataset.

\subsection{Collecting In-the-Wild Multimodal Olfactory Data}
We now describe how we collected the dataset.
\paragraph{Hardware.}
To collect olfactory and visual data in natural environments, we use the natural synchronization between smell and sight.  We chose to use the Cyranose 320 electronic nose~\cite{sensigent_cyranose320}, because it is a  popular handheld sensor that is used in a wide variety of real-world smell sensing applications (see Sec.~\ref{sec:related}). Cyranose consists of a nanocomposite sensor array of 32 sensors. Each sensor responds to different chemical properties of volatile compounds that make up smell, without being specific to one volatile compound. We mount an iPhone 12 camera on Cyranose, angled to view the snout, where the olfactory measurement is collected. Cyranose operates at 2Hz, providing a 32-dimensional olfactory measurement at each timestep. Synchronized with the e-nose, a high-fidelity RGB camera captures the olfactory measurement at $1920\times 1080$ resolution and 15 FPS. 

We record an RGB-D signal using an Intel RealSense D405  (at 15 FPS and $1280\times 720$ resolution), along with  ambient temperature and humidity. We also collect complementary ambient volatile organic compound (VOC) concentrations using a MiniPID2 PPM WR sensor. The PID olfactory measurements reflect the naturally occurring concentrations of smells by diffusion, rather than active sniffing. The e-nose and all sensors are tethered real-time to a mobile station, consisting of battery, data storage, and compute to enable data collection across diverse settings, from parks, apartment settings, to streets. The complete capture set-up is shown in Figure~\ref{fig:capture_setup}.

\begin{figure}[t]
    \centering\includegraphics[width=0.8\linewidth]{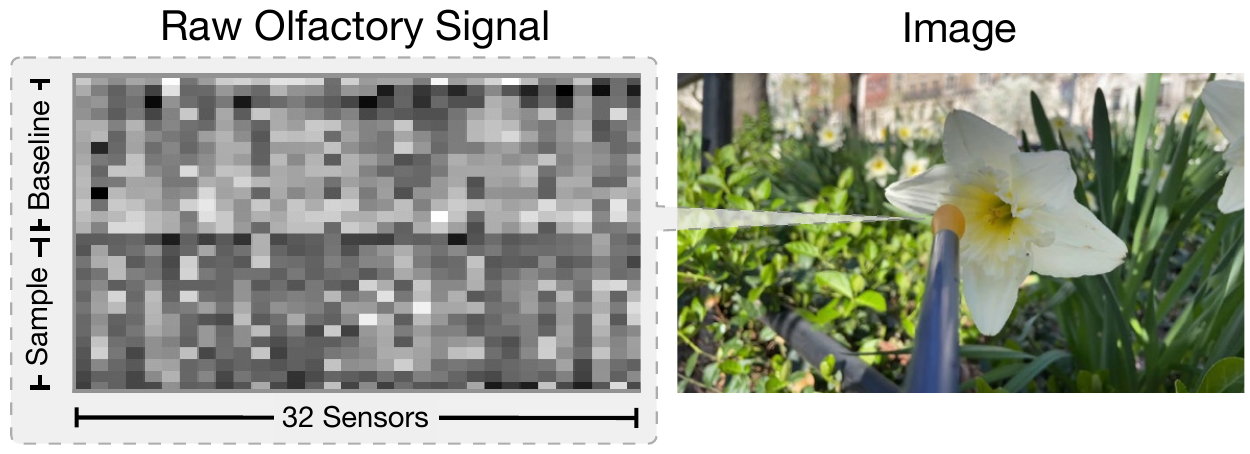}
    \caption{\textbf{Olfactory signal:} The raw smell signal is $T \times 32$ dimensions where $T$ is the capture time. The first part of capture is the baseline phase, where the ambient background smell is sensed. The second part is the sample phase, where the smell of the object of interest is sensed. This example shows the response for a flower.}
    \label{fig:signal}
\end{figure}

The Cyranose obtains a measurement by actively drawing air through its snout and exposing the sampled compounds to an array of 32 sensors. Each sensor is a conductive polymer composite whose electrical resistance changes as odorant molecules are absorbed and cause the polymer to swell. We sample the resistance of all 32 sensors over time as the air sample is acquired, yielding a time-varying response for each channel. The raw olfactory signal is the matrix $x_S \in \mathbb{R}^{T \times 32}$, where each column corresponds to one sensor in the array and each row to one timestep. 

\paragraph{Capturing procedure.}
Sensing objects in the scene requires separating the ambient odors from the odor of the object. For each sample, we first capture the baseline smell of the ambient environment, followed by the smell of the object of interest. Cyranose has two independent air pathways that can ``sniff" outside air into its sensor chamber. The purge/baseline inlet, shown in  Figure~\ref{fig:capture_setup}, on the side of Cyranose, pulls in ambient air, which leaves the sensor array through the exhaust outlet. Through this purge inlet-to-exhaust pathway, we first record the baseline smell for 10 seconds, receiving a ${14 \times 32}$ baseline matrix, representing each sensor for 14 timesteps. During this interval, air is drawn through the side port rather than the measurement inlet to avoid contamination from the target. Next, we record two samples through the main pathway, the snout. For data efficiency, we record two samples for each object from different positions. Both samples are 10 seconds. The raw olfactory data is thus a $28 \times 32$ matrix, which is the concatenation of the baseline and sample stages, as shown in Figure~\ref{fig:signal}.

\begin{SCfigure}[1]%
\centering
\includegraphics[width=0.5\linewidth]{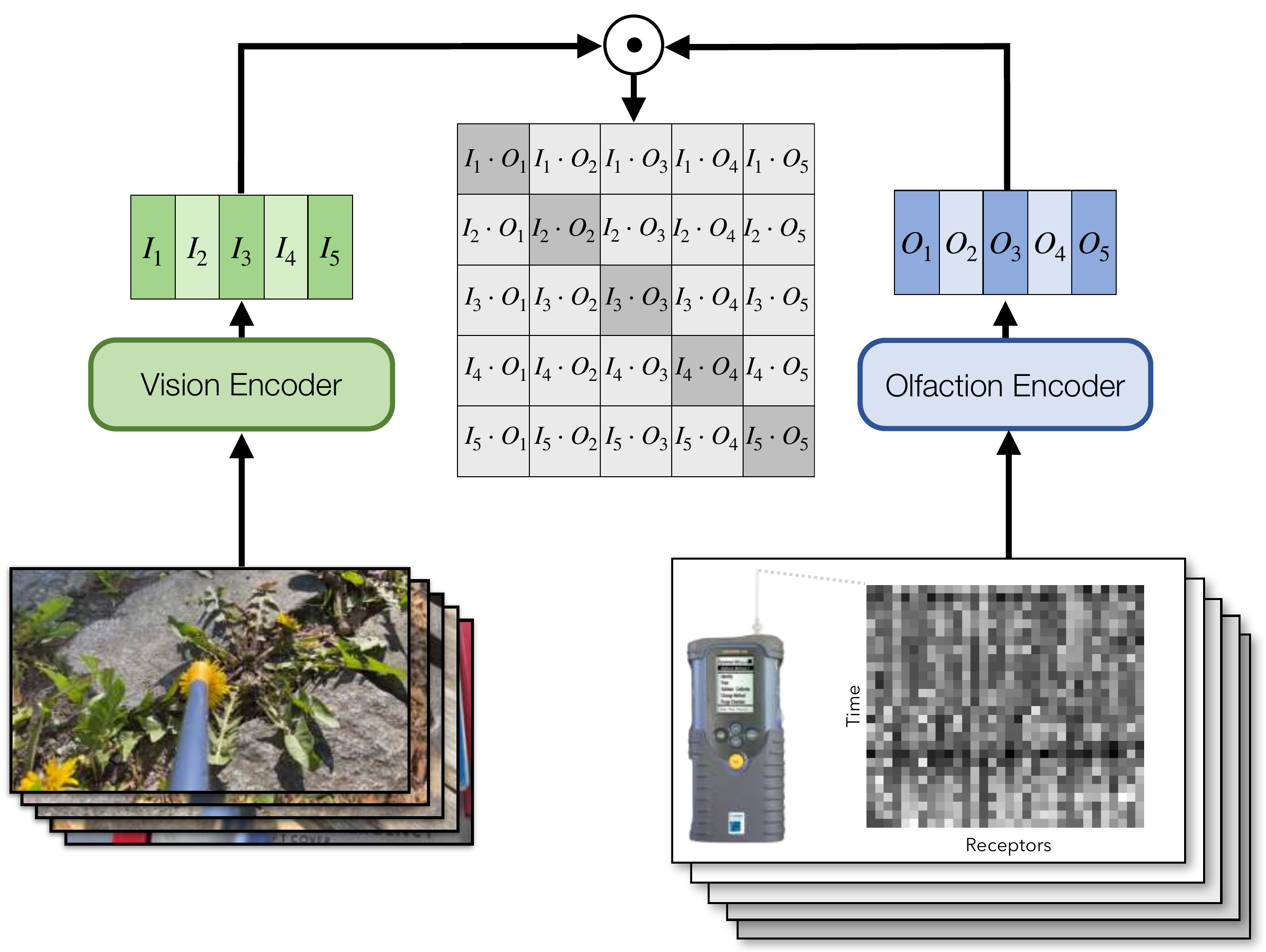}
\caption{{\bf Contrastive olfactory-image learning}. To demonstrate the effectiveness of our dataset, we train general-purpose olfactory representations using contrastive learning. We train the model to align co-occurring visual and smell signals. The visual encoder processes RGB images, while the olfaction encoder processes time-series sensor data from an e-nose.}
\label{fig:method} \vspace{-3mm}
\end{SCfigure}

\paragraph{Labeling the dataset.} \label{dataset:annotation} Figure~\ref{fig:dataset} visualizes the scale and diversity of our olfaction and vision dataset. We used VLMs and the images in our dataset to automatically label the objects and materials. For materials, we used the Matador visual taxonomy of materials \cite{CAVE_427}. Using both views available in our dataset and this taxonomy as a closed set of categories, we generated material labels with VLMs (GPT-4o). For objects, we manually wrote a closed set of 49 categories that spans our dataset, then generated vision labels with VLMs (GPT-4o).  We manually labeled the scene categories for each sample, assigning each data collection session into one of 8 scene categories. To validate the VLM-generated material labels, we manually audited 50 randomly sampled material annotations and observed 84\% agreement with human judgment; the most common failure mode was glass, which the VLM tended to confuse with the material behind a window.

\paragraph{Dataset split.} We uniformly split the dataset into train and validation splits. Since we collect two samples of each object during the capture procedure, we ensure that each sample appears in the same split, thus preventing overlap between the train and validation sets. The dataset has 7K olfactory-vision pairs, 3.5K unlabeled objects, 70 hours of raw video from both cameras, and 196K timesteps of raw smell measurement (baseline and sample stage olfactory measurements). 

This splitting scheme is very simple and avoids the same objects appearing in both splits. However, one limitation is that it leaves potential correlations between examples recorded in the same session. In particular, due to the relatively small number of distinct scenes, scene recognition performance is primarily a measure of a model's ability to determine which smells come from the same scene (rather than predicting the semantic labels from smell). We therefore encourage users of our dataset to consider other splitting strategies if these correlations are an issue.

\paragraph{Dataset analysis.}
Figure~\ref{fig:dataset} shows qualitative examples from the dataset
and Figure~\ref{fig:histogram} shows 
the distribution of materials and objects. We collected data across 60 sessions over two months. 
 The dataset settings include several parks, university buildings, offices, streets, libraries, apartment settings, and dining halls. Each location had multiple data collection sessions. Our dataset has 41\% outdoor and 59\% indoor environments. 

\paragraph{Institutional Review Board (IRB)} Our university's IRB reviewed the collection procedure and determined it to be not human subjects research. No personal identifying information was collected.

\begin{figure}[t]
    \centering
    \includegraphics[width=\linewidth]{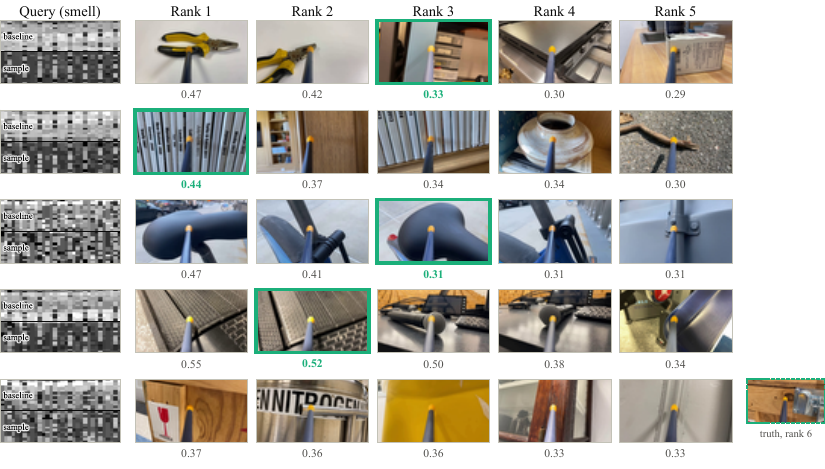}
    \caption{{\bf Cross-modal retrieval qualitative results.} We use our joint embeddings to match smell to images. Given a query smell, we find the images in the dataset that are the closest match in embedding space. Each row shows a reference smell query along with the top 5 image retrievals predicted by our model. The ground-truth smell-image pair is highlighted in green.}
    \label{fig:cross-modal}
\end{figure}

\begin{table}[t]
{\small 
\centering
\setlength{\tabcolsep}{5pt}
\renewcommand{\arraystretch}{1.1}
\begin{tabular}{lccccc}
\toprule
\textbf{Smell Encoder} &
\textbf{Mean Rank} $\downarrow$ & \textbf{Median Rank} $\downarrow$ &
\textbf{R@5} $\uparrow$ & \textbf{R@10} $\uparrow$ & \textbf{R@20} $\uparrow$ \\
\midrule
Chance                  & 467            & 467         & 0.54          & 1.07          & 2.14 \\
\midrule
MLP (Smellprint)        & 375.9          & 329         & 2.04          & 3.43          & 6.22 \\
CNN (Raw Smell)         & 118.4          & 41          & 12.9          & 21.1          & 32.6 \\
MLP (Raw Smell)         & 159.5          & 56          & \textbf{17.3} & 24.2          & 33.8 \\
Transformer (Raw Smell) & \textbf{104.0} & \textbf{28} & 16.5          & \textbf{29.6} & \textbf{43.1} \\
\bottomrule
\end{tabular}}
\vspace{1mm}
\caption{\textbf{Cross-modal retrieval quantitative results.} Recall @ $K$ reported as percentages. We evaluate on $N =933$ test samples.}\vspace{-7mm}
\label{tab:retrieval}
\end{table}

\vspace{-2mm}
\section{Applications of Our Dataset}
\vspace{-1mm}
\label{method}

As an application of our dataset, we use the correspondence between vision and olfaction to learn self-supervised representations for olfaction. The resulting representation can be used for cross-modal retrieval as well as classification tasks.

\vspace{-2mm}
\subsection{Multimodal Contrastive Learning}
\vspace{-1mm}
Humans have a limited sense of smell and there are relatively few words to describe smells compared to other senses \cite{weisgerber1928geruchssinn,yeshurun2010odor} (although see \cite{wnuk2020smell} for possible exceptions). This gap makes it challenging to establish the label
taxonomy and gather annotations on the scale required for effective machine learning. We instead will learn olfactory representations from unlabeled examples, leveraging
cross-modal associations between smell and sight. Inspired by Contrastive Language-Image Pretraining (CLIP), we use contrastive learning to train a joint embedding between smell and images, which we analogously term Contrastive Olfaction-Image Pretraining (COIP).

\begin{figure}[t]
\begin{center}
\includegraphics[width=\linewidth]{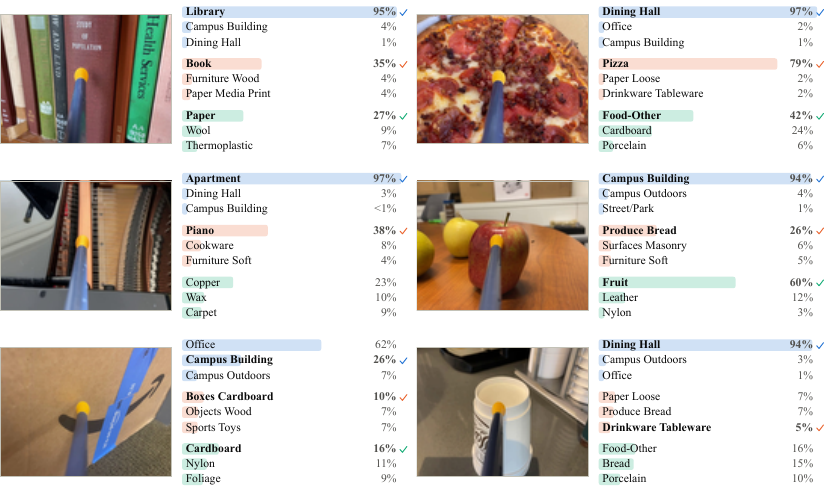}
\end{center}
\vspace{-2mm}
\caption{{\bf Recognizing scenes, objects, and materials from smell.} We show the top 3 predictions from linear probing on the smell encoder. The predictions are from smell alone and the image is shown for visualization purposes only. Predictions are organized by color: \textcolor{figscene}{blue} indicates scene classification, \textcolor{figobject}{orange} indicates object classification and \textcolor{figmaterial}{green} indicates material classification.} %
\label{fig:recognition}
\end{figure}

\begin{table}[t]
\centering
\setlength{\tabcolsep}{4pt}
\renewcommand{\arraystretch}{1.1}
\begin{tabular}{llccccccccc}
\toprule
& & \multicolumn{3}{c}{\textbf{Materials}} & \multicolumn{3}{c}{\textbf{Objects}} & \multicolumn{3}{c}{\textbf{Scenes}} \\
\cmidrule(lr){3-5} \cmidrule(lr){6-8} \cmidrule(lr){9-11}
\textbf{Method} & \textbf{Input} & Scratch & SSL & Rand & Scratch & SSL & Rand & Scratch & SSL & Rand \\
\midrule
Chance & & 1.9 & 1.9 & 1.9 & 2.0 & 2.0 & 2.0 & 12.5 & 12.5 & 12.5 \\
MLP & Smellprint & 3.8 & 2.0 & 6.0 & 3.3 & 5.0 & 5.9 & 42.2 & 32.5 & 31.4 \\
Transformer & Raw smell & 2.3 & 14.0 & 7.1 & 13.8 & 18.4 & 12.3 & 91.0 & 90.4 & 72.7 \\
CNN & Raw smell & 11.9 & 12.3 & 9.4 & 17.9 & 19.8 & 8.7 & 99.5 & 95.0 & 74.5 \\
\bottomrule
\end{tabular}
\vspace{2mm}
\caption{{\bf Recognizing odorants.} We evaluate classification accuracy on our dataset. We use various models and pretraining strategies to recognize scenes, objects, and materials from olfaction alone. For each architecture, we compare end-to-end learning with scratch initialization (Scratch), self-supervised representations with linear probes (SSL), and linear probe with random weights (Rand).}
\label{tab:classification}
\end{table}

Given the dataset of smell and corresponding visual data $\{ \mathbf{x}_S^i, \mathbf{x}_I^i \}_{i=1}^N$, we learn olfactory and visual representations $f_{\theta_S}$ and $f_{\theta_I}$ by jointly training both encoders using a contrastive loss \cite{oord2019representationlearningcontrastivepredictive}:
\begin{equation}
\label{eq:contrastive_loss}
{\small 
\begin{aligned}
\mathcal{L}_{I,S}
&= -\sum_{i=1}^{N}
\log \frac{\exp\left(f_{\theta_I}(\mathbf{x}_I^i) \cdot f_{\theta_S}(\mathbf{x}_S^i) / \tau\right)}
{\sum_{j=1}^{N} \exp\left(f_{\theta_I}(\mathbf{x}_I^i) \cdot f_{\theta_S}(\mathbf{x}_S^j) / \tau\right)},%
\end{aligned}}
\end{equation}
where $\tau = 0.07$ is the temperature. We analogously define the smell to image loss $\mathcal{L}_{S,I}$, where the denominator sums over the visual modality. We minimize both losses to learn the representations $f_{\theta_I}$ and $f_{\theta_S}$: \begin{equation}
\label{eq:combined_loss}
\begin{aligned}
\arg\min_{\theta} \; 
\mathcal{L}_{I,S} +
\mathcal{L}_{S,I}.
\end{aligned}
\end{equation}
By associating sight and smell (Figure~\ref{fig:method}), we learn a representation that can support the downstream interpretation of olfactory stimuli for multiple tasks. 
We apply these learned representations to retrieval and recognition tasks.

\subsection{Input Signals and Architectures}

We experiment with two different input signals in our dataset: a raw representation that has no pre-processing, and a hand-crafted feature space that is widely used in machine olfaction research.  

\paragraph{Raw signal.} Firstly, we directly use the raw signal from the sensor, which is an $T \times 32$ matrix representing the resistance of the 32 sensors inside the Cyranose over $T$ timesteps. We directly input this matrix into the neural network, which then does contrastive learning, and allows end-to-end learning. We experiment with both convolutional neural networks (CNNs) and transformers as the backbone. By learning representations on this raw signal, there is the potential to discover highly powerful representations for olfaction that outperform hand-crafted features.

\paragraph{Smellprint.} Secondly, we compare using a hand-crafted olfactory feature called a smellprint, which is widely used in representing smell from a Cyranose sensor~\cite{dutta2002bacteria,shelley2025impact,visser2020smell,ferreira2023first}. We use it as a baseline for the feature encoding.
The smellprint produces a $32$-dimensional vector from the raw smell matrix, and it summarizes the sensor response to odorants relative to the ambient environment. However, it discards many signals from the raw input, such as the 2nd-order statistics (e.g., correlations between different sensors).

The smellprint is computed by applying Savitzky--Golay filtering (window length $w$, polynomial order $p$) independently to each sensor time series in both the baseline and sample stages. Let $R_{i,j}$ denote the (filtered) resistance of sensor $i\in\{1,\dots,32\}$ at time index $j$. Let $B$ be the baseline indices, and $S = S_1 \cup S_2$ the union of the two sample windows. The per-sensor smellprint feature is the relative response of the sample peak over the ambient baseline:
\begin{equation}
\mathrm{S}_i \;=\; \frac{R_{\max,i} - R_{0,i}}{R_{0,i}},
\quad\text{where}\quad
R_{0,i} = \frac{1}{|B|}\sum_{j \in B} R_{i,j},
\quad
R_{\max,i} = \max_{j \in S} R_{i,j}.
\label{eq:smellprint}
\end{equation}
We feed the $32$ dimensional vector to a multilayer perceptron (MLP) before contrastive learning.

\vspace{-2mm}
\section{Tasks and Experimental Results}
\vspace{-1mm}
Our dataset directly supports two olfactory understanding tasks: cross-modal retrieval between olfaction and vision and recognition tasks including scene, material, and object classification. To further validate the effectiveness of our model, we provide experiments for a fine-grained discrimination task. We evaluate performance with supervised networks, contrastive unsupervised networks with linear probes, as well as the hand-crafted smellprint.

\label{experiments}

\vspace{-2mm}
\subsection{Cross-Modal Retrieval} 
\vspace{-1mm}

We evaluate the ability to retrieve sight from smell and vice versa. 
\paragraph{Setup.}
For each query pair of smell and vision $\{\mathbf{x}_S^q, \mathbf{x}_I^q\}$ in our held-out test set, we sample a distractor set of images $D = \{\mathbf{x}_I^i \}_{i=1}^{N-1}$. We first embed the query pair into the shared olfactory and visual space to get $\{z_S^q, z_I^q\}$, where $z_S^q = f_{\theta_S}(\mathbf{x}_S^q)$ and $z_I^q = f_{\theta_I}(\mathbf{x}_I^q)$. We also embed every image $\mathbf{x}_I^i$ in $D$ into the same olfactory-visual space: $z_i = f_{\theta_I}(\mathbf{x}_I^i)$. %
We sort every image feature $z_i$ by its distance to the query smell feature $z_S^q$. If $z_I^q$ is closest to $z_S^q$, then it will have a rank of 1.  Following \cite{radford2021learningtransferablevisualmodels, tian2020contrastivemultiviewcoding}, we use median rank, mean rank, and recall @ $K$ to measure the percentage of smell queries for which the matching image embedding is ranked in the top $K$ results. 

\paragraph{Results.} Table \ref{tab:retrieval} compares CNN, MLP, and Transformer architectural variants of our olfactory encoder $f_{\theta_I}$ trained on the raw olfactory data as well as the hand-crafted smellprint. Contrastive pretraining using the smellprint performs better than chance in all metrics. However, training the olfactory encoder on the raw olfactory signal leads to significant improvement compared to the smellprint encoder, independent of architecture. This shows the richer information present in the raw olfactory data, unlocking stronger cross-modal associations between sight and smell. We show qualitative results in Figure \ref{fig:cross-modal}. Retrievals from the model often show semantic groupings. The odor of a book retrieves images of other books, the odor of leaves retrieves images of foliage. These results suggest that the learned representation captures meaningful cross-modal structure. Retrievals also group by material properties. For instance, the odor of moss on a concrete bench retrieves images of moss on tree bark and on another bench, while the odor of a wooden stick retrieves images of groundcover and tree bark.

\vspace{-2mm}
\subsection{Object, Material, and Scene Recognition}
\vspace{-1mm}

We train models to recognize object, material, and scene categories from our dataset.

\paragraph{Setup.} We evaluate how well different olfactory models are able to discriminate scenes, objects, and materials from smell alone.  For each task, we compare architectural variants of the smell encoder (MLP, Transformer, CNN) trained via olfactory–visual contrastive learning against the same encoders with random weights, as well as versions trained on smellprint features rather than raw sensory inputs.
We use linear probes on the olfaction representation. To train a probe, we use the activations from the penultimate layer of the olfaction network, and train it to predict the labels derived from the visual stream using GPT-4o on the training set. We then evaluate the probe on the held-out test set.  Linear probes isolate the contribution of the representation itself. %

\paragraph{Results.} As shown in Tab.~\ref{tab:classification}, self-supervised olfaction representations trained with visual supervision outperform baselines. Models trained on raw sensory inputs achieve higher accuracy than models trained with the hand-crafted smellprint features. This demonstrates the strength of training networks on raw smell signals using our dataset. In Figure~\ref{fig:recognition}, we showcase Top 3 predictions from linear probing our smell encoder, spanning diverse scenes, materials, and objects in our test set. As described in Sec.~\ref{dataset}, the scene recognition performance largely reflects the ability to find pairs of odors from the same physical areas, rather than generalization across scenes, which explains the relatively high performance.

\vspace{-2mm}
\subsection{Fine-grained Discrimination}\label{fine_grained_discrimination}
\vspace{-1mm}

We ask whether learned olfactory representations can capture fine-grained differences. In this benchmark, the goal is to distinguish between two grass species recorded at the same campus lawn, where they co-exist. To test this, we collected alternating samples of both grass species across six 30-minute sessions, yielding a balanced dataset of 256 examples. We trained a linear classifier on the features learned through olfactory–visual contrastive learning and evaluated it on a held-out recording session of 42 samples.

\vspace{-2mm}
\begin{wraptable}[12]{R}{0.5\textwidth}
\vspace{-3mm}
\centering
{\small 
\begin{tabular}{llcc}
\toprule
\textbf{Method} & \textbf{Input} & \textbf{Accuracy}\\
\midrule
Chance & & 50.0\\
\midrule
Random weights  & Smellprint & 66.7 \\
Trained from scratch & Smellprint & 85.7 \\
SSL + linear probe & Smellprint & 90.0  \\ %
\midrule
Random weights & Raw smell & 47.6 \\
Trained from scratch & Raw smell & 52.4 \\
\textbf{SSL + linear probe} & Raw smell & \textbf{92.9} \\
\bottomrule
\end{tabular}}
\vspace{-1mm}
\caption{{\bf Fine-grained discrimination}. We evaluate our olfaction models' ability to discriminate between grass species.}\vspace{-5mm}
\label{tab:finegrained}
\end{wraptable}

\paragraph{Results.} Tab.~\ref{tab:finegrained} shows classification accuracy for discriminating the two grass species. %
Training on the raw olfactory sensor signal yields the highest accuracy---exceeding all variants based on smellprints. These results suggest that olfactory--visual learning preserves more fine-grained information than learning with smellprints, and that visual supervision provides a signal for exploiting this information. %

\vspace{-2mm}
\section{Conclusion} \label{sec:conclusion}
\vspace{-1mm}
We present \datasetname, a real-world multimodal dataset of paired visual and olfactory signals collected in natural, in-the-wild environments. We demonstrated that visual data provides effective supervision for learning olfactory representations through contrastive learning, and that models trained on raw olfactory signals substantially outperform traditional hand-crafted features. %

We see our dataset as opening two new research directions. It takes a step toward linking the fields of computer vision to computational olfaction, which have previously been studied separately. We have shown several ways that visual signals can supervise olfaction, such as through self-supervised contrastive learning with static images, but there are many other supervision cues that vision can provide, such as by conveying how objects change over time and 3D space. Our work is also a step toward creating olfactory datasets that can train models in the wild, rather than in lab settings. %

\looseness=-1
\paragraph{Limitations} Our dataset has several limitations. First, it provides data from two e-noses (Cyranose and a PID sensor). We chose these because they are very popular for recognition tasks (sec.~\ref{sec:related}), portable, and provide complementary information. Like all e-noses, there are many chemicals that they do not sense, and representations do not easily transfer from one sensor to another. Second, while the capturing procedure is much more diverse than previous efforts, the data is influenced by a number of factors, including temperature, humidity, and time of day. Finally, the data was only collected in one city and campus. Thus models trained on the data may not generalize to other locations. While there are many positive applications of machine olfaction, it also has potential downsides, such as reducing privacy by revealing health status and enabling better tracking for military applications.

\section*{Acknowledgments} Funding for this research is provided in part by NSF Awards \#2046910 and \#2339071 and the NSF ERC for Smart Streetscapes. We thank Antonio Torralba for the early encouragement. We also thank Max and his large snout for much inspiration.

\bibliographystyle{plainnat}
\bibliography{references}

\clearpage

\appendix

\section{Appendix}

\subsection{Implementation Details}
\label{sec:implementation}

We provide complete implementation details to support reproducibility. We will release code, trained models, data, and the official train/validation splits.

\paragraph{Architectures.}
All olfactory encoders are trained from scratch and produce $128$-dimensional
embeddings, which following standard practice are L2-normalized before the contrastive loss. The raw
olfactory input is the $28 \times 32$ baseline-plus-sample matrix.%

\textit{MLP (raw smell).} The $28 \times 32$ input is flattened to a
$896$-dimensional vector and passed through three blocks of
$\text{Linear} \to \text{ReLU} \to \text{LayerNorm}$ (hidden dimension $128$
throughout), followed by a linear projection to the $128$-dimensional
embedding space.

\textit{CNN (raw smell).} The $28 \times 32$ input is treated as a single-channel
$2$D image and passed through four convolutional blocks with channel widths
$[32, 64, 64, 128]$, each consisting of a $3 \times 3$ convolution
(padding~$1$), BatchNorm, ReLU, and $2 \times 2$ max-pool. The resulting feature
map is reduced to a $128$-dimensional vector by adaptive average pooling, then
linearly projected to the $128$-dimensional embedding space.

\textit{Transformer (raw smell).} The $28 \times 32$ input is treated as a
sequence of $28$ tokens of dimension $d_{\text{model}}=32$. We prepend a
learnable {\sc cls} token, add sinusoidal positional encodings, and apply a
$6$-layer pre-LN Transformer encoder with $4$ attention heads and feed-forward
dimension $d_{\text{ff}}=128$ (dropout $0.1$). The final {\sc cls}
representation is layer-normalized and linearly projected to the
$128$-dimensional embedding space.

\textit{MLP (smellprint).} The $32$-dimensional smellprint vector is passed
through four blocks of $\text{Linear} \to \text{ReLU} \to \text{LayerNorm}$
with hidden dimensions $[64, 64, 128, 128]$, followed by a linear projection
to the $128$-dimensional embedding space.

\paragraph{Retrieval evaluation.} For each smell query in the held-out test set, the distractor set consists of \emph{all} test images ($N{=}933$). We compute cosine similarity between the query smell embedding and every test image embedding, and rank matches exhaustively. Mean rank, median rank, and Recall@$K$ in Table~\ref{tab:retrieval} are computed over this exhaustive ranking.

\paragraph{Linear probing.} For the recognition tasks, we extract activations from the penultimate layer of the frozen olfactory encoder and train a linear classifier with cross-entropy loss on the training split. We report top-1 accuracy on the held-out test split.

\subsection{VLM Prompt for Labeling}

The following Python function is used to label objects using GPT-4o, where images are passed to GPT-4o along with a structured prompt to select the closest matching object category.

\begin{lstlisting}[style=pythonstyle, caption={Object labeling with GPT-4o.}]
def label_gpt_views(image_path1, image_path2, image_path3, image_path4, indexed_labels, labels):
    image_data1 = image_to_base64(image_path1)
    image_data2 = image_to_base64(image_path2)
    image_data3 = image_to_base64(image_path3)
    image_data4 = image_to_base64(image_path4)
    text_prompt = (
        "You are shown four images where a blue sensor probe with a yellow tip is pointing at the same object "
        "from different angles. This is the same target object being analyzed.\n\n"
        "Choose the best matching category label for this object from the list below.\n\n"
        "Respond with the NUMBER corresponding to the best label. Do not invent labels. "
        "If none are perfect, choose the closest match.\n\n"
        "If the image is gray (with white plus sign), choose unlabeled (number 0).\n\n"
        "Category options:\n" +
        "\n".join(indexed_labels) + "\n\n"
        "Respond with only the number. No text."
    )
    response = client.chat.completions.create(
        model="gpt-4o",
        messages=[
            {
                "role": "user",
                "content": [
                    {"type": "text", "text": text_prompt},
                    {"type": "image_url", "image_url": {
                        "url": f"data:image/png;base64,{image_data1}"
                    }},
                    {"type": "image_url", "image_url": {
                        "url": f"data:image/png;base64,{image_data2}"
                    }},
                    {"type": "image_url", "image_url": {
                        "url": f"data:image/png;base64,{image_data3}"
                    }},
                    {"type": "image_url", "image_url": {
                        "url": f"data:image/png;base64,{image_data4}"
                    }},
                ]
            }
        ],
        max_tokens=10,
        temperature=0
    )
    label_number = int(response.choices[0].message.content.strip())
    return label_number
\end{lstlisting}
\newpage

The following Python function sends two image views of the same object to GPT-4o to determine the object's underlying physical material. The prompt includes detailed instructions and examples to avoid visual or semantic biases.
\begin{lstlisting}[style=pythonstyle, caption={Material labeling from two image views using GPT-4o.}]
def label_gpt_views(image_path1, image_path2, indexed_labels):
    image_data1 = image_to_base64(image_path1)
    image_data2 = image_to_base64(image_path2)
    text_prompt = (
        "You are shown two images where a blue sensor probe with a yellow tip is pointing at the same object "
        "from different angles. This is the same target object being analyzed.\n\n"
        "Your task is to identify the object's **main physical material** - not what it contains, what it's shaped like, or what it's used for.\n\n"
        "Choose the most appropriate material from the following list:\n"
        f"{', '.join(indexed_labels)}\n\n"
        "**Label what the whole object is actually made of.** Ignore gloss, color, texture, logos, or symbolic cues.\n\n"
        "**Examples:**\n"
        "- Any food item other than bread -> 'food-other'\n"
        "- A smooth paper cup -> 'paper'\n"
        "- A shiny white bowl -> 'porcelain' or 'thermoplastic' (depends on shape, stiffness, and context)\n"
        "- A juice dispenser labeled 'orange juice' -> 'thermoplastic', not 'fruit'\n"
        "- A brown padded chair seat -> 'leather', not 'terracotta'\n"
        "- A light-colored sidewalk slab -> 'concrete' or 'cement', not 'asphalt'\n\n"
        "**Do not label based on:**\n"
        "- Color (e.g., orange != terracotta; gray != asphalt)\n"
        "- Function (e.g., juice bottle != fruit)\n"
        "- Gloss (e.g., shiny surface != glass)\n"
        "- Shape (e.g., cup shape != plastic)\n"
        "- Logos or printed text\n"
        "- Any object not being directly probed\n\n"
        "If you are absolutely certain it doesn't belong to any of the materials in the list, choose unlabeled. No text."
        "Now return **only the index** of the best matching material from the list as a number, without quotes. Never return any text such as 'I don't know' or 'unlabeled'."
    )
\end{lstlisting}

\clearpage 
\subsection{Additional Examples of Recognition from Smell}

We include additional sampled examples  showcasing the ability of our smell encoder to recognize scenes, objects, and materials from olfactory input alone. These examples further demonstrate the generalization of our model across diverse contexts in the test set and highlight the semantic structure captured by olfactory representations trained with visual supervision. Each prediction is obtained via linear probing and is color-coded by task type: \textcolor{figscene}{blue} for scenes, \textcolor{figobject}{orange} for objects, and \textcolor{figmaterial}{green} for materials.

\begin{figure}[H]
\centering
\includegraphics[width=\linewidth]{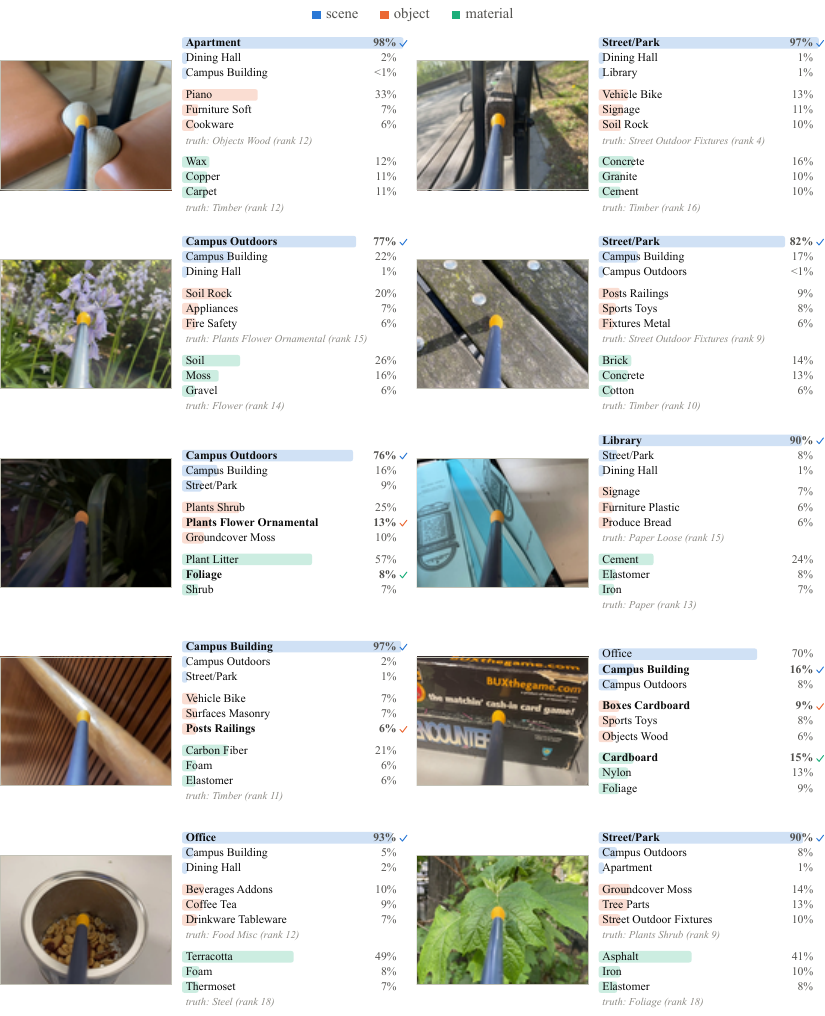}
\caption{Non-cherry-picked examples of recognizing scenes, objects, and materials from
smell (1 of 2). Each tab displays the top-3 predictions obtained via linear probing on the
smell encoder. Prediction types are color-coded: \textcolor{figscene}{blue} for scenes,
\textcolor{figobject}{orange} for objects, and \textcolor{figmaterial}{green} for materials.}
\label{fig:appendix_recognition_1}
\end{figure}

\begin{figure}[H]
\centering
\includegraphics[width=\linewidth]{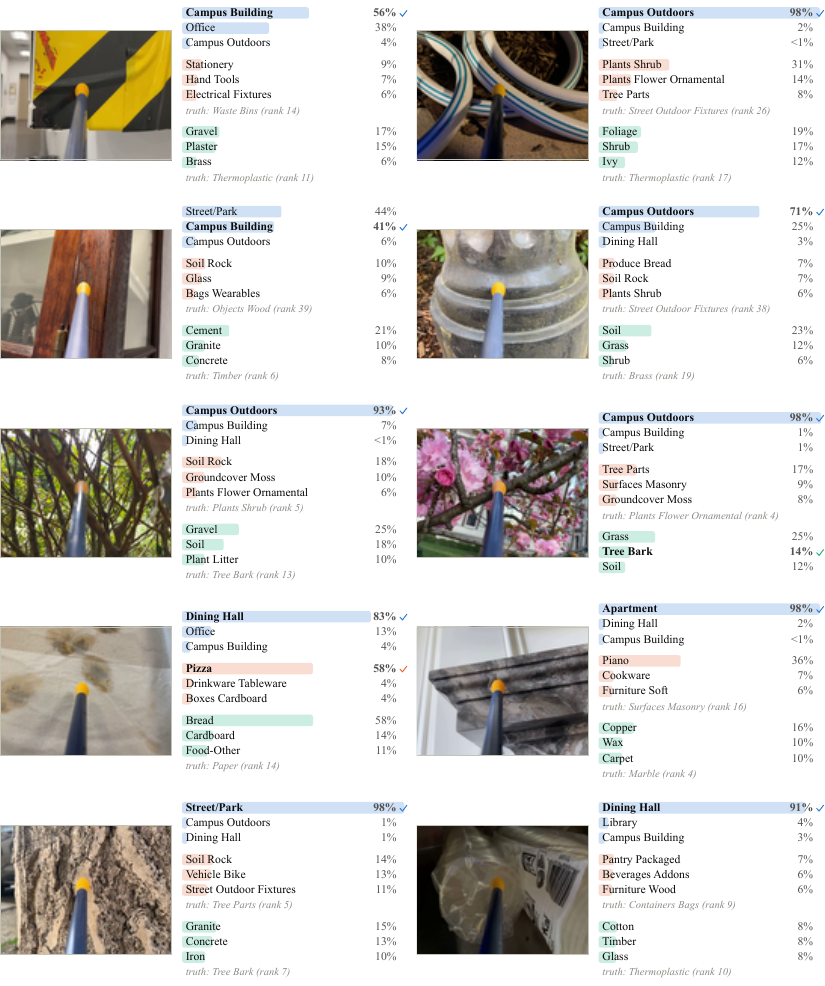}
\caption{Non-cherry-picked examples of recognizing scenes, objects, and materials from
smell (2 of 2). Each tab displays the top-3 predictions obtained via linear probing on the
smell encoder. Prediction types are color-coded: \textcolor{figscene}{blue} for scenes,
\textcolor{figobject}{orange} for objects, and \textcolor{figmaterial}{green} for materials.}
\label{fig:appendix_recognition_2}
\end{figure}

\subsection{Additional Examples of Cross-Modal Retrieval}

We include additional sampled examples of smell-to-image retrieval. These are drawn at random from the test gallery, with no filtering on whether the ground-truth image is retrieved, so the proportion of successful retrievals here reflects the recall reported in Table~\ref{tab:retrieval} rather than a selection. Each row shows a reference smell query along with the top 5 image retrievals predicted by our model. The ground-truth smell-image pair is highlighted in green where it falls in the top 5.

\begin{figure}[H]
\centering
\includegraphics[width=\linewidth]{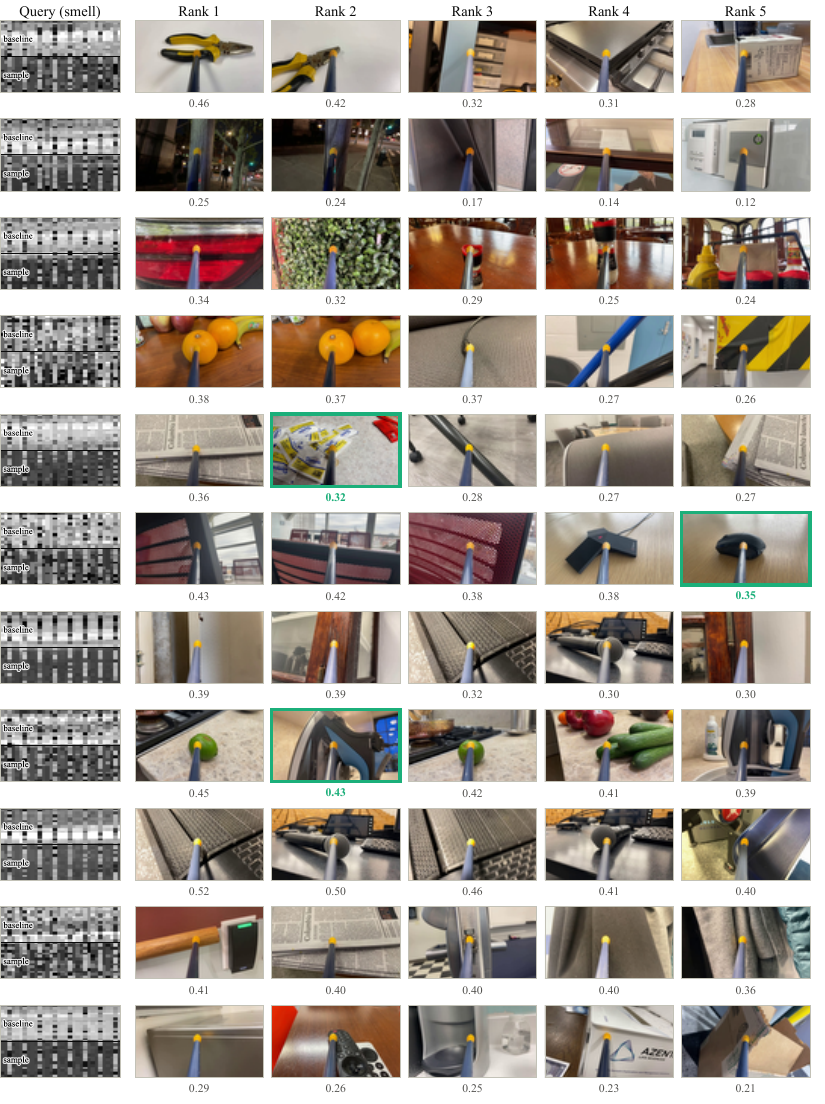}
\caption{Random sampling of cross-modal retrieval (1 of 2). Each row shows a
reference smell query along with the top 5 image retrievals predicted by our model. The
ground-truth smell-image pair is highlighted in green where it falls in the top 5.}
\label{fig:appendix_retrieval_1}
\end{figure}

\begin{figure}[H]
\centering
\includegraphics[width=\linewidth]{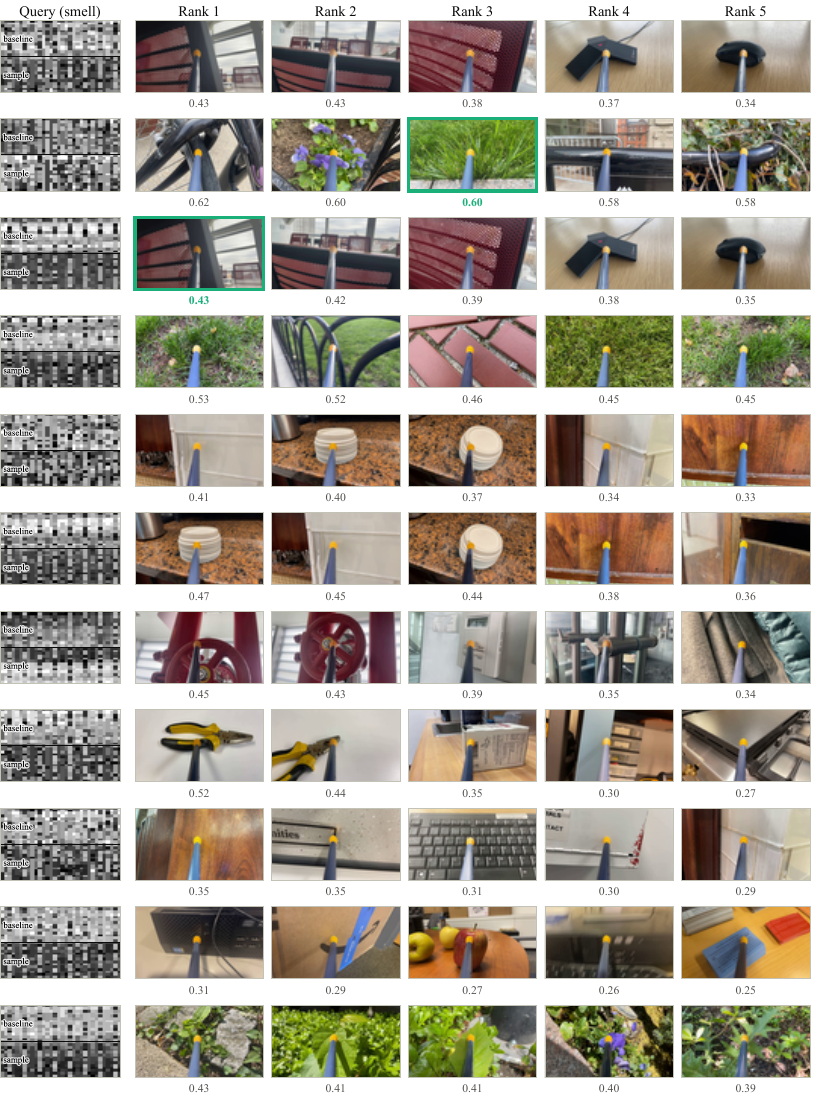}
\caption{Random sampling of cross-modal retrieval (2 of 2). Each row shows a
reference smell query along with the top 5 image retrievals predicted by our model. The
ground-truth smell-image pair is highlighted in green where it falls in the top 5.}
\label{fig:appendix_retrieval_2}
\end{figure}

\end{document}